\documentclass{article} 
\usepackage{iclr2025_conference,times}

\usepackage{amsmath,amsfonts,bm}

\def\eqref#1{equation~\ref{#1}}

\def\1{\bm{1}}

\DeclareMathAlphabet{\mathsfit}{\encodingdefault}{\sfdefault}{m}{sl}
\SetMathAlphabet{\mathsfit}{bold}{\encodingdefault}{\sfdefault}{bx}{n}

\usepackage{hyperref}
\usepackage{url}
\usepackage{graphicx} 
\usepackage[most]{tcolorbox}  
\usepackage{graphicx}
\usepackage{svg}
\usepackage{float} 
\usepackage{booktabs}

\title{Hilbert Neural Operator: Operator Learning in the Analytic Signal Domain}

\author{Saman Pordanesh \\
University of Calgary\\
Department of Software Engineering \\
\texttt{saman.pordaneshasl@ucalgary.ca} \\
\And
Pejman Shahsavari \\
University of Calgary\\
Department of Geoscience \\
\texttt{pejman.shahsavari@ucalgary.ca} \\
\And
Hossein Ghadjari \\
University of Calgary\\
Department of Physics and Astronomy \\
\texttt{hossein.ghadjari@ucalgary.ca} \\
}

\iclrfinalcopy 

\begin{document}

\maketitle

\begin{abstract}
Neural operators have emerged as a powerful, data-driven paradigm for learning solution operators of partial differential equations (PDEs). State-of-the-art architectures, such as the Fourier Neural Operator (FNO), have achieved remarkable success by performing convolutions in the frequency domain, making them highly effective for a wide range of problems. 
However, this method has some limitations, including the periodicity assumption of the Fourier transform. In addition, there are other methods of analysing a signal, beyond phase and amplitude perspective, and provide us with other useful information to learn an effective network. We introduce the \textbf{Hilbert Neural Operator (HNO)}, a new neural operators architecture to address some advantages by incorporating a strong inductive bias from signal processing. HNO operates by first mapping the input signal to its analytic representation via the Hilbert transform, thereby making instantaneous amplitude and phase information explicit features for the learning process. The core learnable operation---a spectral convolution---is then applied to this Hilbert-transformed representation. We hypothesize that this architecture enables HNO to model operators more effectively for causal, phase-sensitive, and non-stationary systems. We formalize the HNO architecture and provide the theoretical motivation for its design, rooted in analytic signal theory.
\end{abstract}

\section{Introduction}
\label{sec:introduction}
Finding a solution for an operator learning a partial differential equation (PDE) is a core challenge in science and engineering. Traditionally, numerical solvers have accurately computed solutions for PDEs in many-query contexts, such as uncertainty quantification, optimal control, and inverse problems. In recent years, deep learning has shown promising alternatives through the framework of neural operators, which learn mappings directly between infinite-dimensional function spaces \citep{lu2019deeponet, kovachki2023neural}. The main advantage of this approach is the ability to learn a single surrogate model that can quickly solve any instance of a given PDE family, thereby reducing the high cost of offline training. 

One of the most successful architectures is the Fourier Neural Operator (FNO) \citep{li2020fourier}. FNO leverages the convolution theorem and the speed of the Fast Fourier Transform (FFT) to parameterize the integral kernel of the operator in the frequency domain. By executing global convolutions as element-wise products in Fourier space, FNO has demonstrated itself to be a robust, discretization-invariant method for various benchmark problems, including viscous fluid dynamics and Darcy flow.

Despite its success, the basic assumption of FNO is hardly tied to the properties of its underlying Fourier basis. Fourier series are optimal for representing periodic, stationary signals; but can be inefficient for problems including some localized features, sharp discontinuities (like shocks), or non-stationary dynamics where frequency content evolves over time. All of these can lead to phenomena such as Gibbs oscillations near shocks, the desire to smooth out sharp features, and limitations noted in this article, which have motivated variants designed to improve local feature representation \citep{gupta2021factorized, li2022fourier}. 

Besides the limitations mentioned for classic neural operators like FNO, there are other motivations for exploring the capabilities of different signal processing tools beyond the Fourier transform. In signal processing, a well-established approach for handling phase information, as well as non-stationary behavior, is to employ the analytic signal via the Hilbert transform. This signal, constructed from two parts—instantaneous phase and amplitude —features through the Hilbert transform, and is a powerful tool widely used in science and engineering for analyzing oscillatory and non-stationary data \cite{rosenblum2021real}.   

To address all mentioned challenges and promises, we propose \textbf{Hilbert Neural Operator (HNO)}, a new architecture that strongly incorporates physics-motivated inductive bias from signal-processing. The main idea of HNO is to learn an operator not only in the domain of the signal itself, but also in the domain of \textbf{analytic signal} \citep{gabor1946theory}. This is achieved by using the Hilbert transform, a linear operator that isolates the signal's instantaneous amplitude and phase \citep{king2009hilbert}. Our main hypothesis is that for many physical systems—mainly those involving wave propagation, causal dynamics \citep{toll1956causality}, or non-stationarity—both instantaneous attributes are helpful fundamentals for approximating the underlying operator and simplifying the learning task for the neural network. 

The contributions of this work are threefold:
\begin{enumerate}
\item We propose and formalize the Hilbert Neural Operator (HNO), a new neural operator architecture that performs spectral convolution on Hilbert-transformed features.
\item We derive the architecture from the first principles of analytic signal theory, providing a clear mathematical justification for its design.
\item We propose an evaluation strategy that identifies specific classes of problems, use cases, and benchmarks where HNO is theoretically promising to perform better. 

\end{enumerate}

\section{The Hilbert Neural Operator}
\label{sec:methodology}

We first briefly review the mathematical background of a Hilbert transform, then formally define the architecture of the Hilbert Neural Operator. 

\subsection{The Hilbert Transform and Analytic Signals}

The Hilbert transform of a real-valued function $v(t)$, denoted $\hat{v}(t)$ or $\mathcal{H}\{v\}(t)$, which is the linear operator defined by the singular integral convolution of $v(t)$ with the kernel $h(t) = 1/(\pi t)$ \citep{bochner1959lectures}:
\begin{equation}
\hat{v}(t) = \mathcal{H}\{v\}(t) = \frac{1}{\pi} \text{p.v.} \int_{-\infty}^{\infty} \frac{v(\tau)}{t - \tau} d\tau
\label{eq:hilbert_def}
\end{equation}
Where p.v. denotes the Cauchy principal value. In the frequency domain, the Hilbert transform, for positive frequencies, corresponds to a phase shift of $-\pi/2$, and for negative frequencies to $+\pi/2$. This is equivalent to multiplying the Fourier transform of the signal, $V(\omega) = \mathcal{F}\{v\}(\omega)$, by $-i \cdot \text{sgn}(\omega)$ \citep{king2009hilbert}:
\begin{equation}
\mathcal{F}\{\hat{v}\}(\omega) = -i \cdot \text{sgn}(\omega) V(\omega)
\label{eq:hilbert_freq}
\end{equation}

The Hilbert transform builds up the construction of \textbf{analytic signal} $v_A(t)$, a complex-valued function whose imaginary part is the Hilbert transform of its real part \citep{gabor1946theory}:
\begin{equation}
v_A(t) = v(t) + i \hat{v}(t)
\label{eq:analytic_signal}
\end{equation}

From the analytic signal, we can uniquely determine the signal's instantaneous amplitude (envelope) $A(t) = |v_A(t)|$ and instantaneous phase $\phi(t) = \arg(v_A(t))$. This decomposition is the main motivation for HNO. It allows the operator to access and learn from these fundamental, time-varying properties directly. A key property of the transform is that applying it twice negates the original function, $\mathcal{H}\{\mathcal{H}\{v\}\}(t) = -v(t)$, which implies the inverse transform is simply $\mathcal{H}^{-1} = -\mathcal{H}$.

\subsection{Architecture}

Similar to the classical FNO, the Hilbert Neural Operator is built as a sequence of operator layers that map the input between function spaces. The key difference lies within the structure of each layer. An FNO layer updates an input function (or feature map) $v_k(x)$ to $v_{k+1}(x)$ as follows:
\begin{equation}
v_{k+1}(x) = \sigma \left( W v_k(x) + \mathcal{F}^{-1} \left( R_\phi \cdot (\mathcal{F}v_k) \right)(x) \right)
\label{eq:fno_layer}
\end{equation}
Here, $\mathcal{F}$ is the Fourier transform, $R_\phi$ is a learnable linear transformation (the spectral kernel) parameterized by $\phi$, and $W$ is a local linear transformation (e.g., a 1x1 convolution).

The HNO layer modifies this structure by performing the spectral convolution on the Hilbert-transformed features. The update rule for an HNO layer is:
\begin{equation}
v_{k+1}(x) = \sigma \left( W v_k(x) + \mathcal{H}^{-1} \left( \mathcal{F}^{-1} \left( R_\phi \cdot (\mathcal{F} (\mathcal{H}v_k)) \right) \right)(x) \right)
\label{eq:hno_layer}
\end{equation}

The operational flow of the spectral path in the HNO layer (the second term inside the activation $\sigma$) is as follows:
\begin{enumerate}
    \item \textbf{Hilbert Transform:} The input function $v_k(x)$ is transformed into its Hilbert representation, $\hat{v}_k(x) = \mathcal{H}\{v_k\}(x)$.
    \item \textbf{Fourier Transform:} The Hilbert-transformed signal $\hat{v}_k(x)$ is mapped to the frequency domain: $\mathcal{F}\{\hat{v}_k\}(\omega)$.
    \item \textbf{Spectral Convolution:} The learnable kernel $R_\phi$ is applied via element-wise multiplication in the frequency domain, typically on a truncated set of low-frequency modes. This is the core learning step, where the model learns patterns from the spectrum of the \textit{analytic features}, not the raw signal.
    \item \textbf{Inverse Fourier Transform:} The result is transformed back to the Hilbert-feature domain.
    \item \textbf{Inverse Hilbert Transform:} The final result is transformed back to the original real-valued signal domain using the inverse Hilbert transform, $\mathcal{H}^{-1} = -\mathcal{H}$.
\end{enumerate}

\begin{figure}[H]
    \centering
    \includegraphics[width=1.0\textwidth]{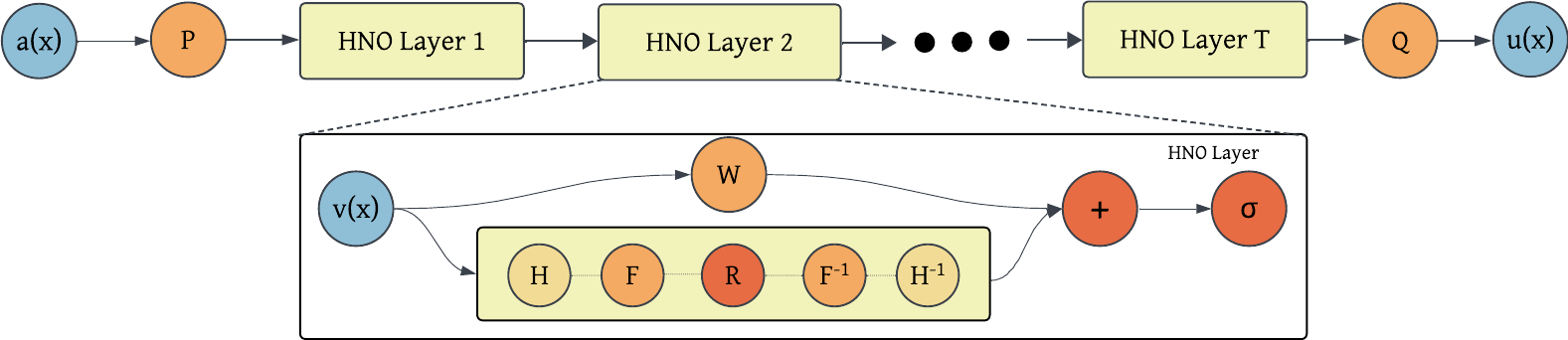}
    \caption{\footnotesize \textbf{The network architecture of the Hilbert Neural Operator}: It first lifts the input signal \(a(x)\) to a high-dimensional latent space using a linear projection, \(P\). The argument \(T\) represents the number of HNO layers. Within each layer, the signal is processed by two parallel branches: (1) a local branch, which uses a standard convolution to capture short-range dependencies, and (2) a global branch, which employs our Hilbert spectral convolution. This global path is designed to learn the operator by extracting features based on the signal's instantaneous phase and amplitude, for capturing long-range interactions. The outputs of these two branches are then combined via summation, followed by a non-linear activation function. After the final HNO layer, a projection layer, \(Q\), maps the latent representation back to the desired output dimension, retrieving the final output \(u(x)\).}
    \label{fig:your_label}
\end{figure}

In this architecture, we force the learned kernel $R_\phi$ to operate on a representation where phase and amplitude dynamics are explicitly defined. As a result, HNO learns the dynamics directly. This provides a strong inductive bias, which is beneficial for the specific problem classes mentioned in the introduction and can give HNO distinct capabilities for a set of use cases. For practical implementation, the Hilbert and Fourier transforms are computed efficiently using FFT-based algorithms, similar to the methods employed in FNO \citep{li2020fourier}. We didn't see a contrast to utilize the benefits of Fourier transform convolutional features within HNO kernel, and leverage its fast multiplication operation in Fourier space. It also still allows us to benefit from the mode truncation advantages, which were introduced at FNO, here in the Hilbert-based operator.

\section{Experiment}
\label{sec:experiment}

\begin{tcolorbox}[colframe=gray!60, colback=gray!5, left=4pt, right=4pt, boxrule=0.5pt]
These are preliminary experimental cases. Research is still ongoing, and more problems may be added to this section.
\end{tcolorbox}

In our experimental design, we test the HNO with three PDE simulations mentioned and benchmarked in the FNO paper \cite{li2020fourier}, including Burgers' equation, Darcy-Forchheimer, and Navier-Stokes equations. All is well described and outlined in the original paper, and we just mention some basics about each problem for the record. We also introduce an additional problem to this experimental design, which is the Lorenz-63 system, based on the dataset presented in Laplace Neural Operator \cite{cao2024laplace}. 

\subsection{Burgers' Equation}
The Burgers' equation is a non-linear PDE with different applications in simulating the flow of a viscous fluid. This equation tries to model fluid mechanics as a simplified prototype of the Navier-Stokes equations. 
\begin{equation}
\partial_t u(x,t) + \partial_x \left( \frac{u^2(x,t)}{2} \right) = \nu \partial_{xx} u(x,t), 
\quad x \in (0,1), \ t \in (0,1]
\end{equation}
\begin{equation*}
u(x,0) = u_0(x), \quad x \in (0,1)
\end{equation*}

\subsection{Darcy Flow}
The second equation, Darcy Flow, is a linear PDE that governs fluid flow. This flow primarily describes the slow, viscous movement of a fluid through a porous medium under the influence of a pressure gradient. 
\begin{equation}
- \nabla \cdot \left( a(x) \nabla u(x) \right) = f(x), 
\quad x \in (0,1)^2
\end{equation}
\begin{equation*}
u(x) = 0, \quad x \in \partial(0,1)^2
\end{equation*}

\subsection{Navier-Stokes Equation}
Navier-Stokes addresses the incompressible equations, which govern the motion of viscous fluid substances and are fundamental in fluid dynamics. The nonlinearity and pressure-velocity coupling make this equation difficult to solve, and they are typically used to model a wide range of phenomena, including laminar flow and turbulence. 
\begin{equation}
\partial_t w(x,t) + u(x,t) \cdot \nabla w(x,t) = \nu \Delta w(x,t) + f(x),
\quad x \in (0,1)^2, \ t \in (0,T]
\end{equation}
\begin{equation*}
\nabla \cdot u(x,t) = 0, 
\quad x \in (0,1)^2, \ t \in [0,T]
\end{equation*}
\begin{equation*}
w(x,0) = w_0(x), \quad x \in (0,1)^2
\end{equation*}

\subsection{Lorenz-63}
The Lorenz system, which serves as an example of three ODEs set, is a mathematical model that simplifies many practical problems, such as electric circuits, atmospheric convection, and forward osmosis. These three coupled ordinary differential equations are known for exhibiting chaotic behaviour from deterministic rules. 
\begin{equation}
\begin{aligned}
\dot{x} &= \sigma (y - x), \\
\dot{y} &= x (\rho - z) - y, \\
\dot{z} &= xy - \beta z - f(t),
\end{aligned}
\end{equation}

\section{Results}
\label{sec:results}

Here we can see the preliminary results on five different cases based on the 4 mentioned problems in Section\ref {sec:experiment}. We tested the HNO versus FNO and LNO (in some cases) through three 1-dimensional and two 2-dimensional examples. The validation error through training epochs is shown in Figure \ ref {fig:results-all}, and the final validation L2 error can be found in Table\ref{tab:results}. 
Datasets that were used in these experiments are already cited in Section\ref{sec:experiment}, and the code that was used to replicate these experiments via the new HNO kernel is stored at the project's GitHub repo: \url{https://github.com/sinapordanesh/XHNO}

\begin{figure}[htbp]
  \centering
  \setlength{\tabcolsep}{2pt}
  \renewcommand{\arraystretch}{0.8}
  \begin{tabular}{@{}ccc@{}}
    \includegraphics[width=0.31\textwidth]{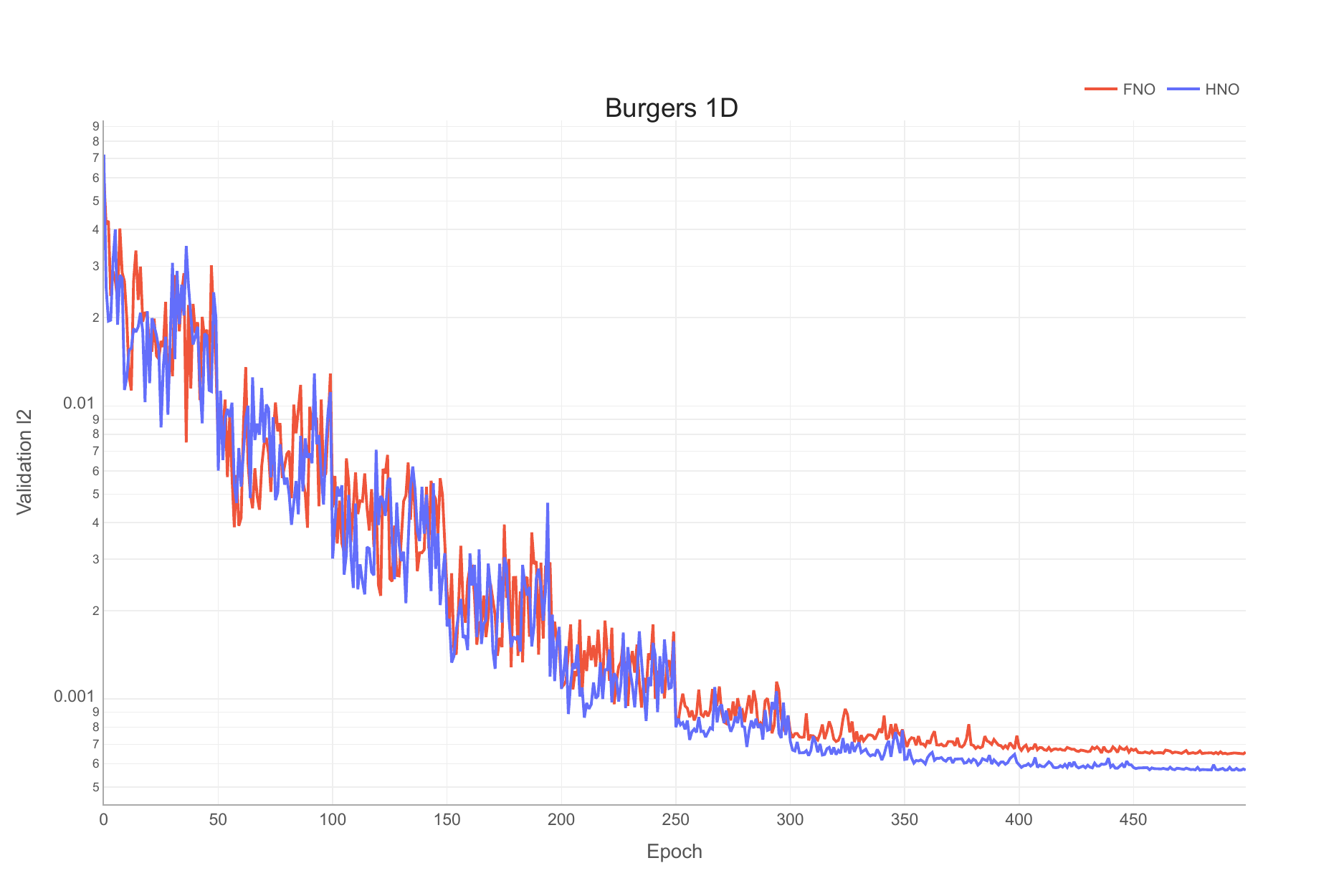} &
    \includegraphics[width=0.31\textwidth]{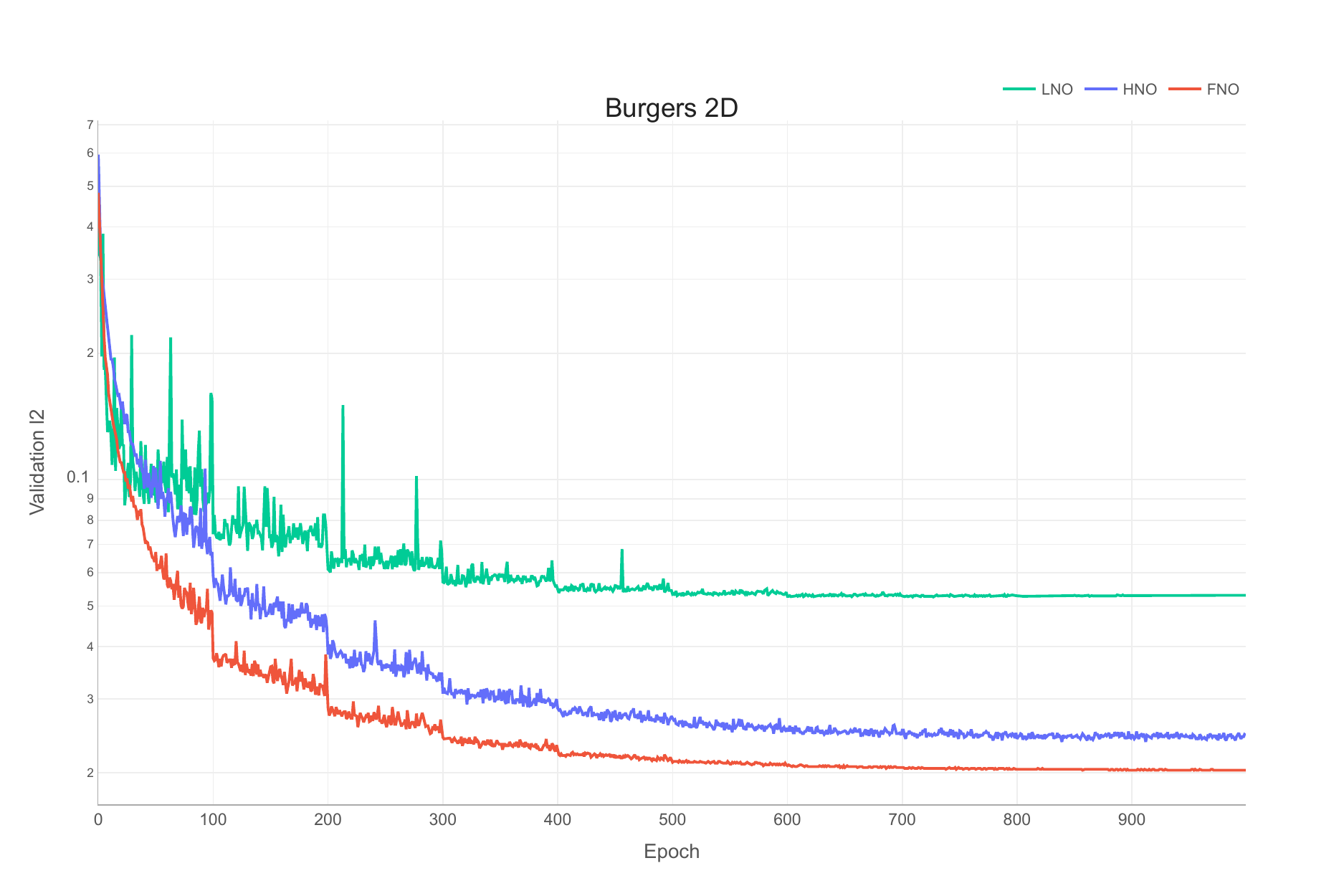} &
    \includegraphics[width=0.31\textwidth]{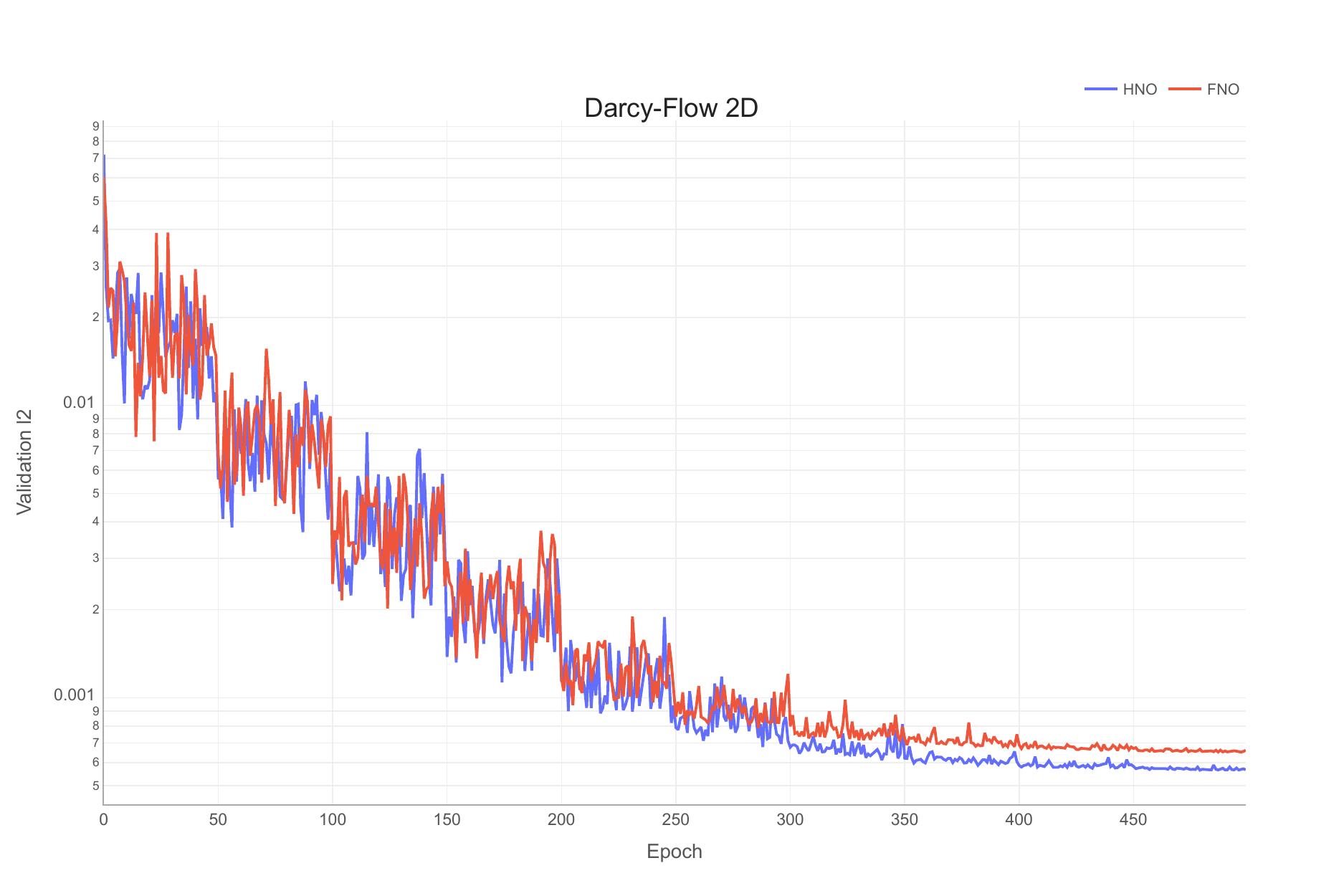} \\[0.5em]
    \multicolumn{1}{c}{\includegraphics[width=0.31\textwidth]{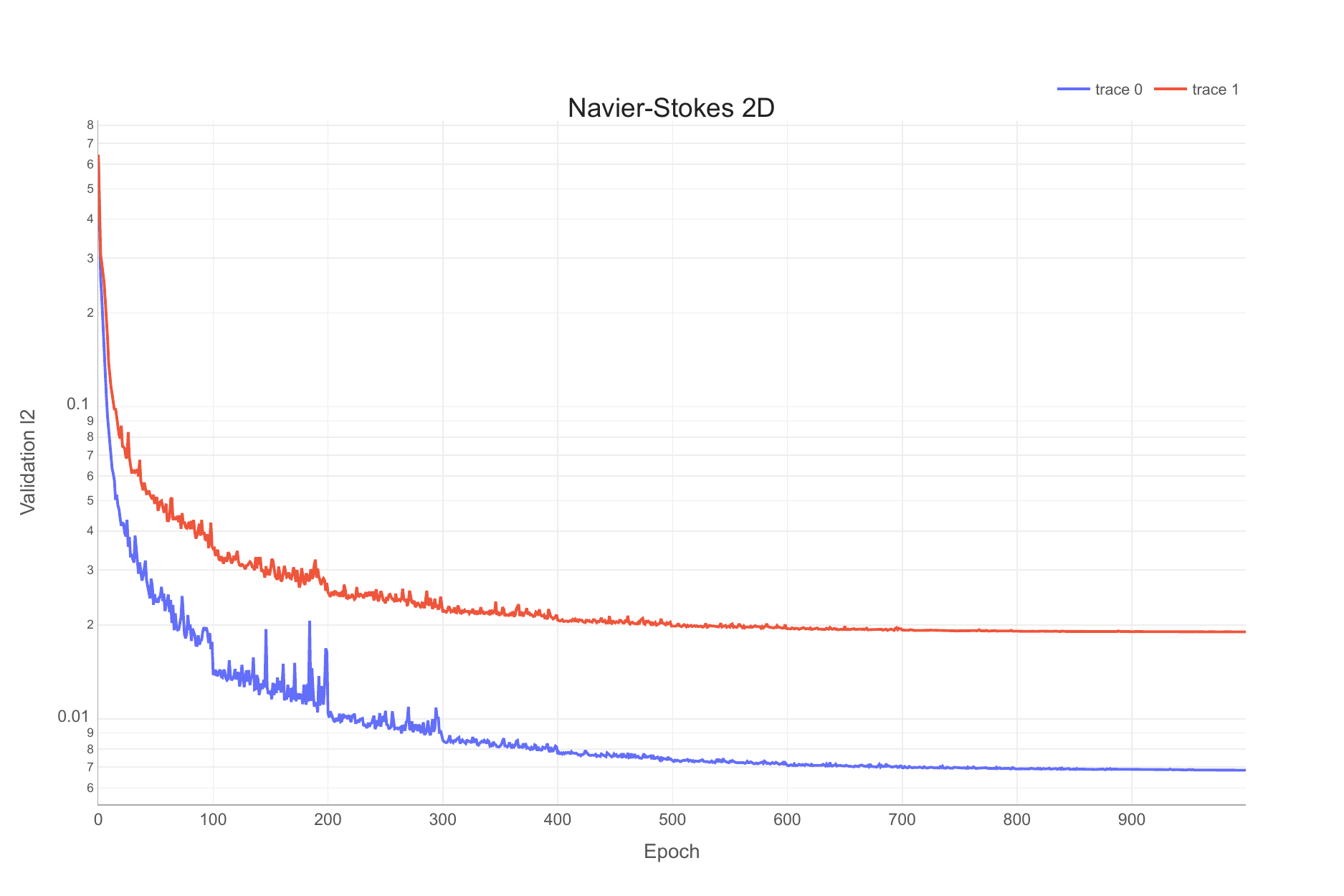}} &
    \multicolumn{1}{c}{\includegraphics[width=0.31\textwidth]{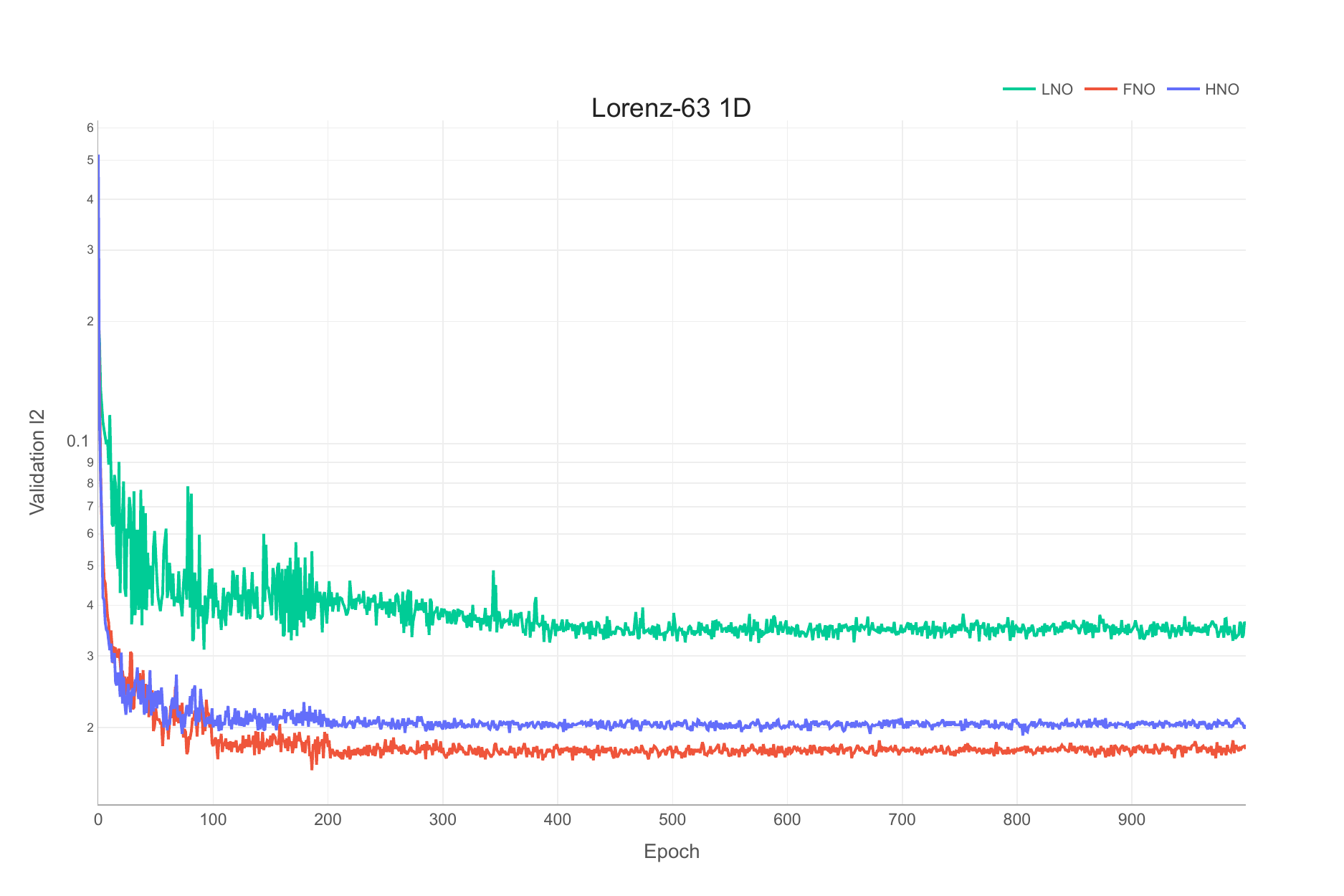}} &
    \multicolumn{1}{c}{}
  \end{tabular}
  
  \vspace{-0.5em} 
  \caption{Comparison of results across five PDE systems: (top row) 1D Burgers, 2D Burgers, and 2D Darcy; (bottom row) 2D Navier-Stokes and 1D L63 equations.}
  \label{fig:results-all}
\end{figure}

\begin{table}[htbp]
  \centering
  \caption{Comparison of test errors across different methods and PDE systems.}
  \label{tab:results}
  \begin{tabular}{@{}lccc@{}}
    \toprule
    \textbf{PDE} & \textbf{HNO} & \textbf{FNO} & \textbf{LNO} \\
    \midrule
    \textbf{Burgers-1D} & $\mathbf{5.72 \times 10^{-4}}$ & $6.57 \times 10^{-4}$ & --- \\
    \textbf{Lorenz63-1D} & $5.06 \times 10^{-3}$ & $\mathbf{3.94 \times 10^{-3}}$ & $2.03 \times 10^{-2}$ \\
    \textbf{Burgers-2D} & $6.03 \times 10^{-5}$ & $\mathbf{4.11 \times 10^{-5}}$ & $1.18 \times 10^{-4}$ \\
    \textbf{DarcyFlow-2D} & $3.83 \times 10^{-2}$ & $\mathbf{1.64 \times 10^{-2}}$ & --- \\
    \textbf{NavierStokes-2D} & $1.90 \times 10^{-2}$ & $\mathbf{6.84 \times 10^{-3}}$ & --- \\
    \bottomrule
  \end{tabular}
\end{table}

Most of the results are showing a better performance for FNO, and with a comparable difference, HNO in the second place. All selected experimental cases originate from well-known and main benchmarks in the scientific machine learning literature, which may not be suitable for presenting the benefits of a special case like HNO. For the next stage of our experiments, we are aiming to discover more tailored examples and use-cases for HNO, like Aucostic wave PDEs, where the Hilbert transform is mostly used to interpret signals. This may require more expert knowledge in this field, and this is where we are calling for collaboration to improve this method, as well as design better experiments. 

\section{Discussion}
\label{sec:disscussion}

In this article, we aim to follow the footsteps of classical neural operators and introduce a new kernel that may offer benefits in certain problems and simulations over other operator learning methods. We are leveraging the analytical signal of the input in a Hilbert-Fourier space to learn some features from the signal that a Hilbert transform provides, such as instantaneous phase and amplitude. The Hilbert transform is mostly used in acoustic signal and sound processing, and other contributors could leverage this method in relevant fields and report the results. 
We are still in the exploratory phase, investigating other use cases beyond the known and mentioned baselines, where this new kernel could perform better or more meaningfully than other approaches, such as FNO \cite{li2020fourier}, LNO \cite{cao2024laplace}, and WNO \cite{tripura2022wavelet}. 

\begin{tcolorbox}[colframe=gray!60, colback=gray!5, left=4pt, right=4pt, boxrule=0.5pt]
This article is primarily aimed at showcasing our interest in this field and calling for collaboration from anyone with ideas to share regarding applications of the HNO in other fields. 
\end{tcolorbox}

\bibliography{references}
\bibliographystyle{iclr2025_conference}

\appendix

\end{document}